\documentclass[letterpaper]{article} 
\usepackage{aaai25}  
\usepackage{times}  
\usepackage{helvet}  
\usepackage{courier}  
\usepackage[hyphens]{url}  
\usepackage{graphicx} 
\usepackage{natbib}  
\usepackage{caption} 
\usepackage{algorithm}
\usepackage{algorithmic}

\usepackage{amsmath,amssymb,amsthm,subfig,mathtools}
\usepackage{autobreak}
\usepackage{graphicx}
\usepackage{cleveref}
\usepackage{multirow}
\usepackage{tikz}
\usepackage{booktabs}
\usepackage{enumitem}
\usepackage{afterpage}
\usepackage{placeins}
\usepackage{graphicx}
\usepackage[toc,page]{appendix}
\usepackage{xparse}

\usepackage{newfloat}
\usepackage{listings}
\DeclareCaptionStyle{ruled}{labelfont=normalfont,labelsep=colon,strut=off} 
\floatstyle{ruled}
\newfloat{listing}{tb}{lst}{}
\floatname{listing}{Listing}
\nocopyright
\title{Teaching Models to Improve on Tape}

\author {
    Liat Bezalel\textsuperscript{\rm 1},
    Eyal Orgad\textsuperscript{\rm 1},
    Amir Globerson\textsuperscript{\rm 1,\rm 2}
}
\affiliations {
    \textsuperscript{\rm 1}Tel Aviv University\\
    \textsuperscript{\rm 2}Google Research\\
    liatbezalel@mail.tau.ac.il, eyalorgad@mail.tau.ac.il, 
    amirg@google.com
}

\usepackage{bibentry}

\begin{document}

\maketitle

\begin{abstract}
Large Language Models (LLMs) often struggle when prompted to generate content under specific constraints. However, in such cases it is often easy to check whether these constraints are satisfied or violated. Recent works have shown that LLMs can benefit from such ``corrective feedback''. Here we claim that this skill of LLMs can be significantly enhanced via training. We introduce an RL framework for teaching models to use such rewards, by simulating interaction sessions, and rewarding the model according to its ability to satisfy the constraints.
We refer to our method as CORGI (Controlled Generation with RL for Guided Interaction), and evaluate it on a variety of controlled generation tasks using unlabeled training data. We find that CORGI consistently outperforms the baseline reinforcement learning method that does not incorporate conversational feedback. Furthermore, CORGI's interactive framework enables meta-learning, allowing the LLM to generalize better to guided interaction in new tasks. Our results clearly show that conversational optimization, when combined with reinforcement learning, significantly improves the effectiveness of LLMs in controlled generation contexts.
\end{abstract}

%

\section{Introduction}

Large Language Models (LLMs) have become quite effective at following instructions from users. Despite this progress there are still many tasks where LLMs will not respond correctly. A particular subclass of cases is those where it is relatively easy to judge whether a response is correct or not. For example, if the task is to generate a sentence with six words, it is fairly easy for to check if this constraint is violated. 

A natural question is whether such feedback can be used in order to make the LLM produce a correct response. This exciting prospect has been considered in several works recently, including \citet{NEURIPS2023_91edff07} and 
\citet{shinn2024reflexion}. Indeed, these works show that  feedback can be ``written to the LLM context'', and the LLM can use this to improve its generation. Furthermore, it has been shown that LLMs can learn to utilize external tools \citep{NEURIPS2023_d842425e, NEURIPS2023_871ed095}, thereby enhancing their results and reliability.

Building on the potential of feedback, we aim to explore whether LLMs can use this signal more effectively. We focus on controlled generation tasks, where LLMs often struggle \citep{sun-etal-2023-evaluating}. 
Specifically, we consider the important subclass of tasks where it is easy to check if the generated text satisfies the desired control requirements (e.g., the ``generate sentence with six words" example above''). We hypothesize that LLMs can be trained to more effectively use corrective feedback in these cases, and we approach this goal through Reinforcement Learning (RL).

\begin{figure*}[t]
  \includegraphics[width=\linewidth]{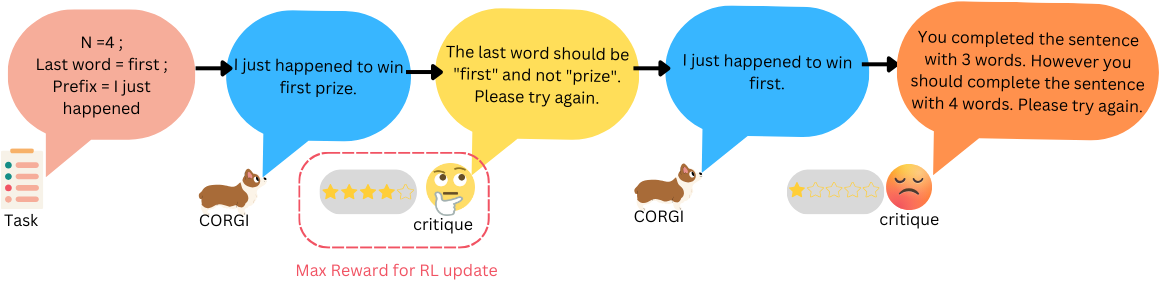}
  \caption{ An example of the CORGI setup. We consider a dialogue between a generator and a critique. Here the generator is tasked with completing a given sentence in precisely four words, with the final word being ``first''. The critique evaluates the responses of the generator, providing both feedback and a score (illustrated as a star rating). The LLM receives a reward based on the highest score assigned by the critique throughout the dialogue history. To prioritize improvement on the more challenging length constraint, we set the constraint weights to 80\% for length and 20\% for the last word constraint.}
  \label{fig:teaser}
\end{figure*}

We implement an RL based approach for training LLMs to improve their corrective skills. To do so, our training data consists of several controlled generation tasks. During training we let the model interact with corrective feedback over multiple iterations, and provide a reward corresponding to the provided feedback. Specifically, the delivered reward corresponds to the best feedback obtained in the interaction session. To understand why this is the reward chosen, consider a controlled generation task, where the corrective feedback says how many constraints are satisfied. Then, one can always choose the output that satisfied the most constraints. See Figure \ref{fig:teaser} for an illustration of an interaction session and corresponding reward.

For the above reward scheme, we train the model using RL, to obtain a model that can better use corrective signals. Our first observation is that on the tasks that the model was trained on, there is a significant improvement in performance, as compared to baselines that are not trained to use feedback. An even more surprising finding is that these trained models also improve when using feedback in tasks they were not trained on. This supports our conjecture that LLMs can improve on the ``meta-skill'' of using corrective feedback. 

Our results demonstrate the significant potential of teaching LLMs to effectively incorporate feedback into their output generation processes. By demonstrating improvements in controlled generation tasks, we highlight how LLMs can learn to refine their outputs based on corrective input, leading to more accurate and reliable results. This suggests that the ability to utilize feedback is not merely a task-specific enhancement but may represent a more generalizable skill that would demonstrate gains across a wide range of tasks.

\section{CORGI Setup and Model}
This section outlines our proposed approach: Controlled Generation with RL for Guided Interaction, or CORGI.

\subsection{Problem Definition}
In controlled generation tasks \citep{see-etal-2019-makes, gehman-etal-2020-realtoxicityprompts, sun-etal-2023-evaluating} the model is presented with specific constraints or requirements that define the desired characteristics of the generated text. For these tasks we assume access to an automated critique environment. Our primary aim is to teach an LLM to effectively incorporate critique feedback, while bypassing the need for labeled data. We aim to train a model capable of generating text that adheres to these constraints, even in scenarios where the model has not been directly trained on the task and its critique system.

Formally, consider a set of $M$ controlled-generation tasks $\{
\mathcal{T}_i\}_{i=1}^M$. For each task ${\mathcal{T}_j}$, let $D_{\mathcal{T}_j} = \{x_i\}^{N_j}_{i=1}$ denote an unlabeled dataset of controlled generation prompts, where $x \in \mathcal{X}$ is the task input. Additionally, for each task, let $C_{\mathcal{T}_j}$ denote the automated critique. This critique is defined as a function $C_{\mathcal{T}}: \mathcal{X} \bigtimes \mathcal{Y} \rightarrow \mathbb{R} \bigtimes \mathcal{Z}_{\mathcal{T}}$ which takes as an input the task's input and a generated output $y \in \mathcal{Y}$, and returns a score $r \in \mathbb{R}$ and a textual feedback $z \in \mathcal{Z_\mathcal{T}}$. For example, in Figure \ref{fig:teaser} the critique offers textual feedback, notifying the generator that it used the wrong last word or that the output was too long. This feedback is accompanied by a score, illustrated as a star rating in the figure. 

Our objective is to train a model on the $M$ available task datasets using their critique environment. Importantly, we would like for the model to also perform well on new tasks $\mathcal{T}'$ that have their own critique feedback $C_{\mathcal{T'}}$ and are not used for training.

In what follows, we first describe the iterative use of critiques, and the corresponding reward the model receives, and then describe how this reward can be optimized using RL.

\subsection{Iterative Use of Critiques} 
We consider the setting where the model dynamically receives feedback \cite{shinn2024reflexion, yao2022react, NEURIPS2023_91edff07}. In this framework, the model receives real-time textual feedback from the critique, allowing it to adjust and refine its output. The key aspect of our setting is that the critique is invoked every time the model produces a new response, so that the model gets multiple attempts to produce a good answer. We limit the number of interactions to $K$. After $K$ iterations (or sooner if the model successfully completes the task), the model returns the response that received the highest score from the critique. 

\paragraph{Reward Definition} 
Our key goal in this work is to train LLMs to improve their performance when using critiques in an iterative fashion, and we use an RL approach towards this end.
The key component in such a training approach is the reward that the model receives in each session. In our setting it is clear that this reward is simply the best critique score obtained in the session, since this is the quality of the returned output.\footnote{We note that the response with best critique score need not be the last one. For example in a length constrained generation task where the goal is to product 6 words, it's possible that the model had three tries and returned lengths of 2,5,3 in these. In this case, the second attempt will be the one returned.} 
Formally, in each session, the generator produces up to $K$ outputs, with each output $o_i$ being evaluated by the critique and assigned a score $s_i$. The session's reward is then defined as:

\begin{equation}
    r = \max\limits_{i=0,\ldots, K} s_i
\end{equation}

\subsection{Using RL to Improve the Use of Feedback}
To maximize the expected reward across sessions, we utilize reinforcement learning. Specifically, we employ Proximal Policy Optimization (PPO) \citep{schulman2017proximal}, a method proven effective in Reinforcement Learning from Human Feedback (RLHF) \cite{ouyang2022training}. Building on previous applications of reinforcement learning in large language models \cite{ouyang2022training, achiam2023gpt}, we define the actions $a_i \in \mathcal{A}$ as the tokens selected from a finite vocabulary $\mathcal{V}$ by the policy, while the state corresponds to the prompt presented to the model at each time step.

We further integrate reinforcement learning into the CORGI framework, enhancing its ability to adapt and optimize based on real-time feedback. We designed a generalized environment simulating a conversation with an automated critique for reinforcement learning purposes. The trajectory is defined by the interactions generated by the policy with the automatic critique.  Formally, a chat between the LLM and the critique is represented as $(x, a_{0_0}, ..., a_{0_{n_1}}, z_1, a_{1_1}, ..., z_{t}, a_{t_0}, ..., a_{t_{n_t}})$ where $x$ is the task specific prompt, $\{z_i\}_{i=1}^t$ is the textual feedback from a critique $\mathcal{C}$. The actions $\{a_{i_j}\}_{j=1}^{n_i}$ correspond to the tokens generated by the LLM in the $i^{th}$ attempt to generate the output defined in the prompt.\footnote{For example, if the prompt is ``Generate a sentence with 6 words then $a_{3_\cdot}$ would be the third try of generating this sentence, after two previous tries that received a critique feedback.} An example for a chat can be seen in Figure \ref{fig:teaser}. Each episode begins with sampling a prompt $x$ from the dataset and concludes either when the output receives a perfect score from the critique or after $K$ outputs. Each output is constrained by a maximum of $T$ tokens or terminates upon generating the end-of-sentence (EOS) token.

Our goal is then to maximize the PPO objective function:
\begin{equation}
\begin{split}
J(\theta) & = \mathbb{E}[\min(r(\theta)\hat{A}_{\theta_{\text{ref}}}(s,a), \\ & \text{clip}(r(\theta), 1-\epsilon, 1+\epsilon)\hat{A}_{\theta_{\text{ref}}}(s,a))] ~,
\end{split}
\end{equation}
where $r(\theta) = \frac{\pi_{\theta}(a|s)}{\pi_{\theta_{\text{ref}}}(a|s)}$ is defined as the probability ratio between the new policy and the reference policy, $\hat{A}_{\theta_{\text{ref}}}(s,a) = \hat{Q}(s,a) - \hat{V}(s)$ is the estimated advantage function using a value head, and $\epsilon$ is a hyperparameter. To differentiate between action tokens (i.e., the model outputs) and non-action tokens (i.e., the task prompt and critique feedback), we apply a mask to the non-action tokens. Specifically, the reward and value functions for these non-action tokens are set to zero.

In this way, the CORGI framework utilizes both textual feedback and the critique's scoring, teaching the policy to adjust to the textual feedback in real-time and enhancing the reward signal by incorporating the feedback text into the state.

\begin{figure*}[t]
  \includegraphics[width=0.75\linewidth]{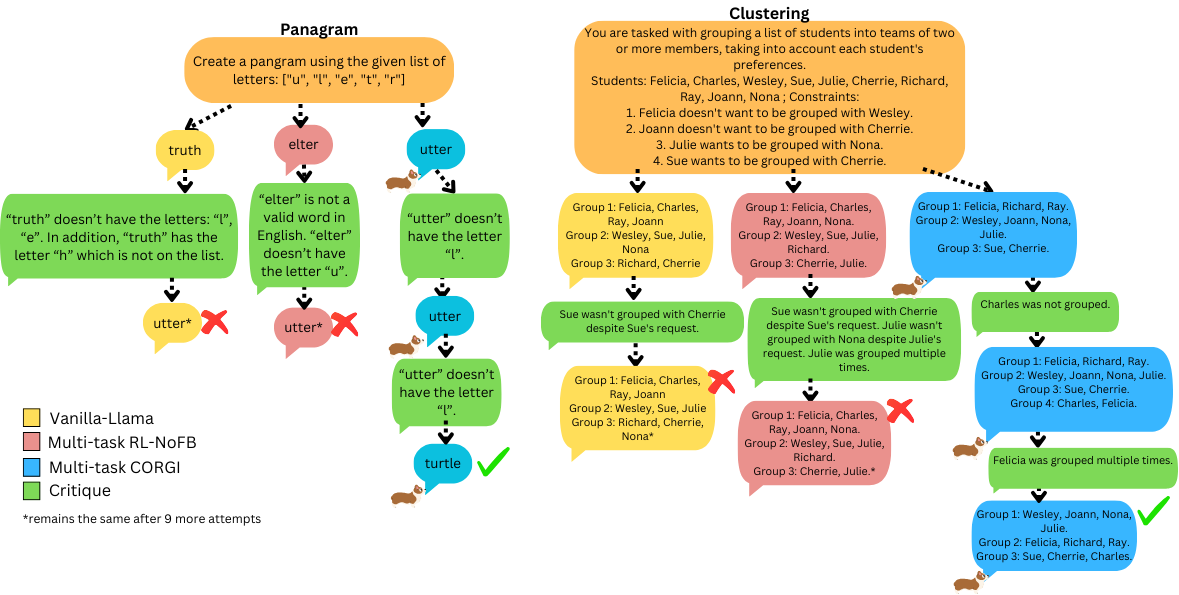}
  \centering
  \caption{Meta-learning examples. The figure shows responses of three models on two prompts: one for the Panagram task and one for the Clustering task. The three models are Vanilla Llama, RL-NoFB and CORGI. The latter two are trained on multiple source-tasks jointly.
  It can be seen that in these instances CORGI arrives at a correct output after several iterations, whereas the other models get stuck repeating a suboptimal solution.}
  \label{fig:corgi_mtl_example}
\end{figure*}

\section{Experiments}
In what follows we apply our CORGI approach to several controlled generation problems. 

\subsection{Data}
\label{sec:data-section}
We conduct experiments across eleven diverse tasks. We include five tasks inspired by those introduced by \citet{sun-etal-2023-evaluating}: Sentiment Reviews Generation, Story Generation, Rationale Generation, Numerical Planning, and Controlled Paraphrase Generation. Due to the unavailability of the original data and code, we reconstructed the dataset. 

We further evaluated our approach on a broader set of tasks, including Style Transfer \citep{pryzant2020automatically}, CommonGen-lite \citep{lin-etal-2020-commongen}, Program Synthesis (Numeric category) \citep{srivastava2023beyond}, MBPP (sanitized subset) \cite{austin2021program}, and CommonGen-Hard \cite{NEURIPS2023_91edff07}. Additionally, we introduced two new tasks: Panagram and Clustering, described below. 

In the Panagram task, the model is required to generate a word using a given list of letters. The generated word must include all specified letters, with each letter used at least once and possibly multiple times. 

The Clustering task explores a more algorithmic challenge. Here, the input is a list of student names, and the model must group the students into clusters of two or more, taking into account their preferences, namely which students they prefer or do not want to be grouped with. The critique provides as feedback the constraints that were violated.



Further details on data construction and the reward system for all tasks are provided in the Appendix. 

For simplicity, we normalize all rewards to be in the $[0,1]$ range, where $1$ indicates a perfect response. For all these tasks, we include a textual response by the critique, which is provided to the generator whenever the reward is less than the maximal value of one (corresponding to perfect generation).

We categorize the tasks into two groups:

\begin{enumerate}
\item \textit{Source Tasks} - These tasks are used for training with RL. We used the following source tasks: Sentiment Reviews Generation, Story Generation, Rationale Generation, Numerical Planning, and Controlled Paraphrase Generation.
\item \textit{Target Tasks} - These tasks are only used for evaluation and are not used during RL. Namely, we use these to evaluate how well the model transfers to problems it wasn't trained on. We used the following target tasks: Style Transfer, Clustering, Panagram, and CommonGen-lite. Additionally, we introduced a subgroup of Llama-3-specific target tasks, which includes Program Synthesis, MBPP, and CommonGen-Hard. These tasks were specifically evaluated with Llama-3 due to poor performance by Llama-2.
\end{enumerate}

\subsection{Pre-processing} For each task, we created a prompt outlining the constraints and including two few-shot examples. Detailed examples can be found in the Appendix. For each task, we generated 7,500 training prompts and 500 validation prompts. For Rationale Generation, Controlled Paraphrase Generation, Common-Gen lite, Program Synthesis (numeric category), MBPP (sanitized subset), Common-Gen Hard tasks we used the predefined train/dev/test splits - sampling the training prompts and validation prompts from their corresponding splits. For tasks without predefined train/dev/test splits, we additionally produced 1,000 test prompts.

\subsection{Experimental Setup}
\subsection{Baselines} 
Our study utilizes the models Llama-2-7b-chat  \cite{touvron2023llama} and Llama-3-8b-instruct \cite{llama3modelcard} as the initial checkpoints to which RL is applied.\footnote{Since our focus is on RL, we cannot use models that cannot be trained.}  We examine two distinct baseline configurations:
\begin{enumerate}
    \item \textbf{Vanilla-Llama} - The Llama models, without any additional training. This setup aligns with related work \citep{shinn2024reflexion, NEURIPS2023_91edff07} where feedback is derived from an external source.
    \item \textbf{RL-NoFB} -  This model serves as a reinforcement learning baseline without receiving feedback during training. The model is given a controlled generation prompt and generates a response. A critique is applied to this response, and its value is used as a reward for the PPO  algorithm. Unlike CORGI, this model does not receive textual feedback during training nor does it make multiple generation attempts.
\end{enumerate}

During evaluation, we allow the baseline models to make multiple attempts and provide them with the same feedback as CORGI to assess their ability to utilize this feedback. For each model, we select the attempt with the highest score according to the critique. We note that such multiple attempts can only make the baseline stronger. 

\subsection{Training} We trained CORGI and RL-NoFB using Llama-2-7b-chat and Llama-3-8b-instruct with Proximal Policy Optimization (PPO). The CORGI framework was implemented using the TRL \cite{vonwerra2022trl} library, which we also utilized for RL-NoFB by setting the number of attempts to one. The Adam optimizer was employed with a learning rate of $10^{-5}$ and Adaptive KL control \cite{ziegler2019fine}, with initial coefficients set to $0.05$ for Llama-2 and $0.075$ for Llama-3. Due to training instability, the KL coefficient was adjusted to $0.3$ for the rationale generation task in Llama-3. For CORGI training, the number of attempts was limited to four per prompt due to computational constraints. Training was conducted on a single NVIDIA A100-SXM4-80GB GPU, taking 4 days for the multi-task setting and 12-24 hours for the single-task setting.

As mentioned above, we were interested in the ability of the model to transfer to tasks that were not seen during training, and the advantage of training on multiple tasks. Thus, we considered two different training scenarios:

\paragraph{Single-Task Training} In this setting, we trained RL-NoFB and CORGI on the source tasks (Sentiment Reviews Generation, Story Generation, Rationale Generation, Numerical Planning, and Controlled Paraphrase Generation). Each model was trained on only one source-task, resulting in five trained models. 

\paragraph{Multi-Task Training} Here, we perform multi-task training. Namely, we train RL-NoFB and CORGI using data aggregated from all the above five source tasks. Namely, this results in one model trained on all tasks jointly.

\begin{table*}[h]
\centering
\begin{tabular}{llllll}
\hline
Model                                         & Training                     & Setting  & Source Tasks  & Target Tasks  & Overall       \\ \hline
\multicolumn{1}{l|}{\multirow{5}{*}{Llama 2}} & None                         & Few-shot & 68.8          & 55.4          & 62.8          \\ \cline{2-6} 
\multicolumn{1}{l|}{}                         & \multirow{2}{*}{Single-Task} & RL-NoFB  & 71.4          & 53.9          & 63.7          \\
\multicolumn{1}{l|}{}                         &                              & CORGI & \textbf{73.7} & \textbf{54.5} & \textbf{65.2} \\ \cline{2-6} 
\multicolumn{1}{l|}{}                         & \multirow{2}{*}{Multi-Task}  & RL-NoFB  & 70.7          & 52.4          & 62.5          \\
\multicolumn{1}{l|}{}                         &                              & CORGI & \textbf{74.6} & \textbf{58.5} & \textbf{67.5} \\ \hline
\multicolumn{1}{l|}{\multirow{5}{*}{Llama 3}} & None                         & Few-shot & 86.3          & 84.2          & 84.3          \\ \cline{2-6} 
\multicolumn{1}{l|}{}                         & \multirow{2}{*}{Signle-Task} & RL-NoFB  & 87.6          & 83.1          & 85.6          \\
\multicolumn{1}{l|}{}                         &                              & CORGI & \textbf{92.8} & \textbf{84}   & \textbf{88.9} \\ \cline{2-6} 
\multicolumn{1}{l|}{}                         & \multirow{2}{*}{Multi-Task}  & RL-NoFB  & 87.3          & 84            & 85.9          \\
\multicolumn{1}{l|}{}                         &                              & CORGI & \textbf{90.8} & \textbf{85.2} & \textbf{88.3} \\ \hline
\end{tabular}
\caption{Results on controlled generation tasks. The table shows performance on source tasks and target tasks, excluding Llama-3-specific target tasks (results for these tasks reported in Figure \ref{fig:llama3_specific_target_tasks_bar}).
All results are for the best generated response after $10$ attempts for Llama-2 and $20$ attempts for Llama-3. Standard deviation is estimated with bootstrap. The first row for each model corresponds to the Vanilla-Llama model. The second row and third rows (Single-Task) show the average results for five different models trained on a single training task. The fourth and fifth rows (Mutli-Task) show the results of a single model per training setting, each trained on five source tasks simultaneously. We provide the average success rate for the source tasks (column 3), the average success rate for the target tasks (column 4), and the overall average success rate for all nine tasks (column 5). The results per task can be seen in the Appendix}
\label{tab:summarized-results}
\end{table*}

\begin{figure}[h]
  \includegraphics[width=0.8\linewidth]{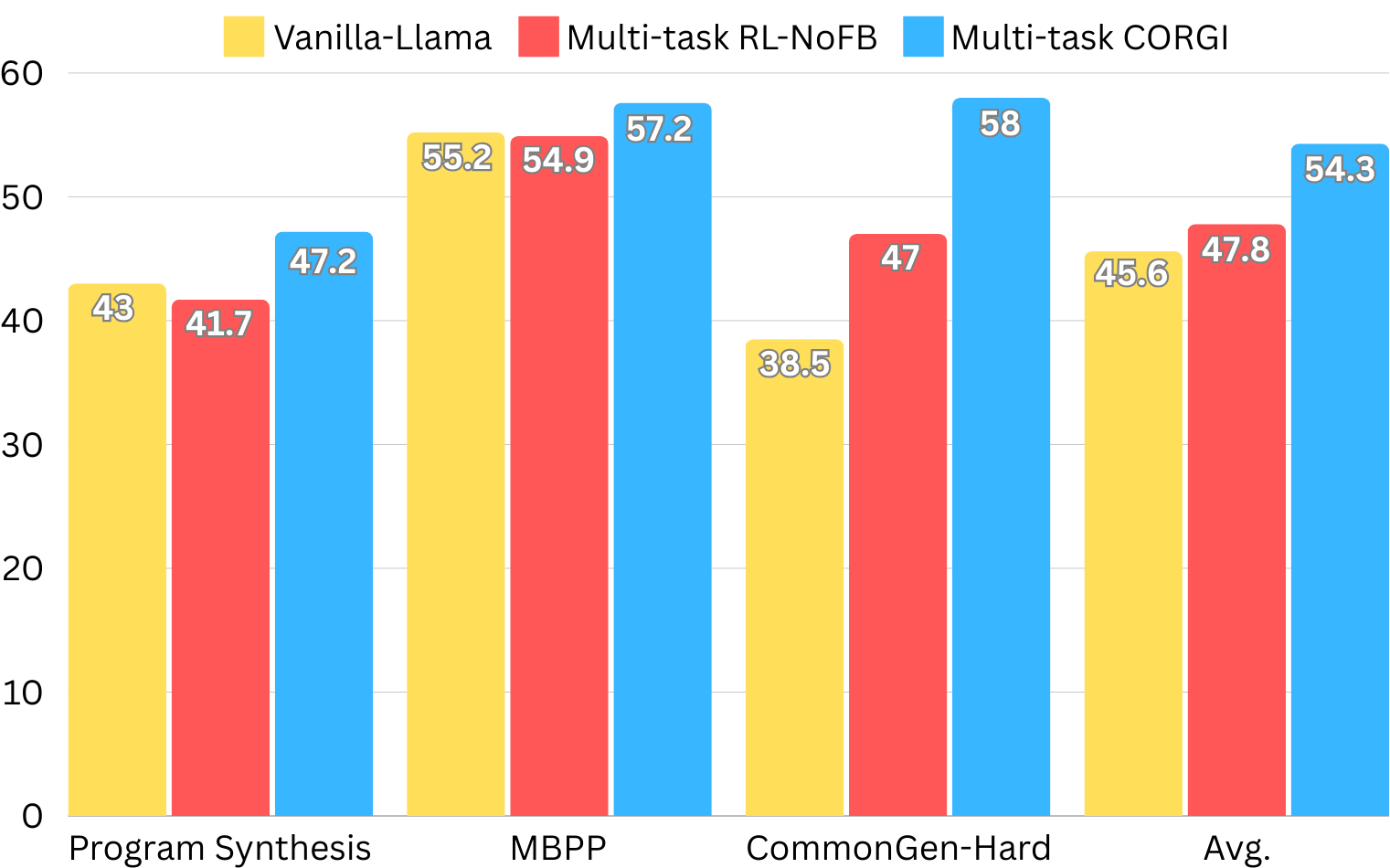}
  \centering
  \caption{ Multi-task Performance on the Llama-3-Specific Target Tasks. The results show that CORGI significantly benefits from transfer learning, outperforming both the Vanilla-Llama and RL-NoFB configurations.}
  \label{fig:llama3_specific_target_tasks_bar}
\end{figure}

\subsection{Evaluation}
We evaluate the aforementioned settings on the eleven tasks. When evaluating on the source tasks, in the single-task setting, we evaluate each model on the task it was trained on.
To assess the meta-learning capabilities of the multi-task models, we include evaluations on the Target Tasks (i.e., Style-Transfer, CommonGen-lite, Clustering, Panagram, and Llama-3 Specific target tasks - Program Synthesis, MBPP, CommonGen-Hard).

During inference, we employ the same automated critique environment used during CORGI training. Each generator, including the baseline models, uses greedy decoding and receives $10$ attempts for Llama-2 and $20$ attempts for Llama-3 \footnote{The increased number of attempts for Llama-3 is due to its larger context size limit of 8192 tokens, compared to Llama-2's 4092 tokens.}, incorporating critique's feedback to complete the tasks.

\begin{figure*}[t]
\centering
\begin{minipage}{.33\textwidth}
  \centering
  \includegraphics[width=1\linewidth]{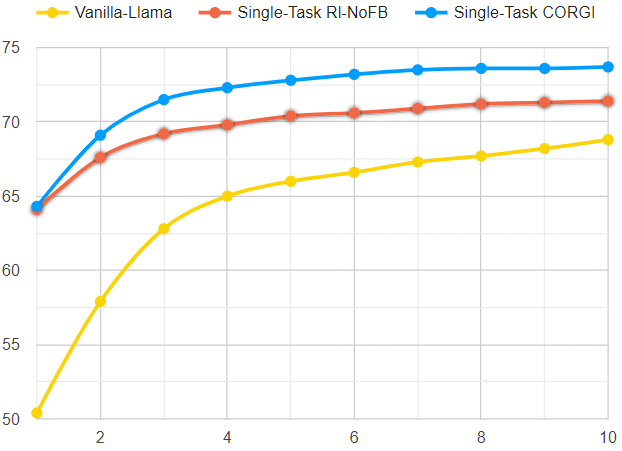}
  \label{fig:seen-tasks-bar}
\end{minipage}%
\begin{minipage}{.33\textwidth}
  \centering
  \includegraphics[width=1\linewidth]{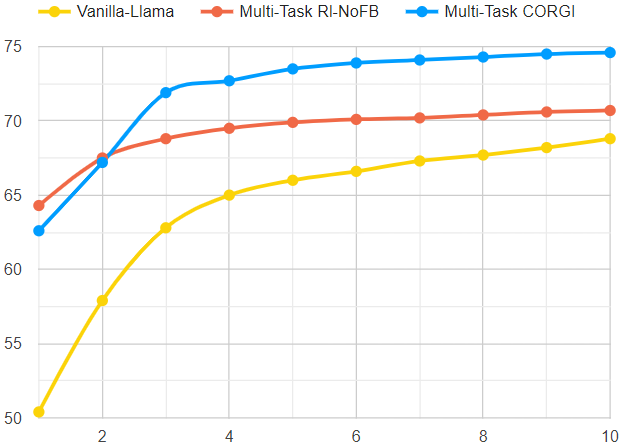}
  \label{fig:unseen-tasks-bar}
\end{minipage}
\begin{minipage}{.33\textwidth}
  \centering
  \includegraphics[width=1\linewidth]{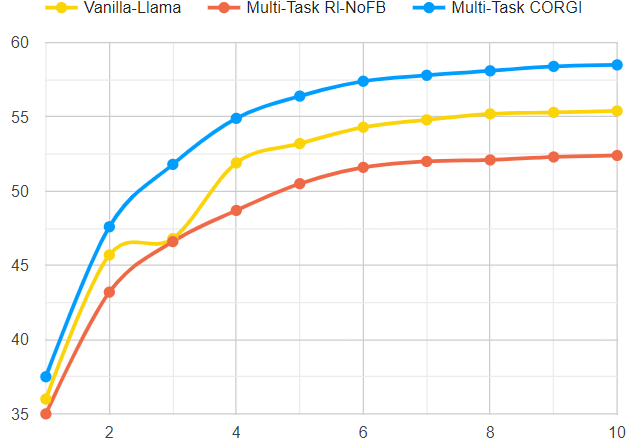}
\end{minipage}
\begin{minipage}{.33\textwidth}
  \centering
  \includegraphics[width=1\linewidth]{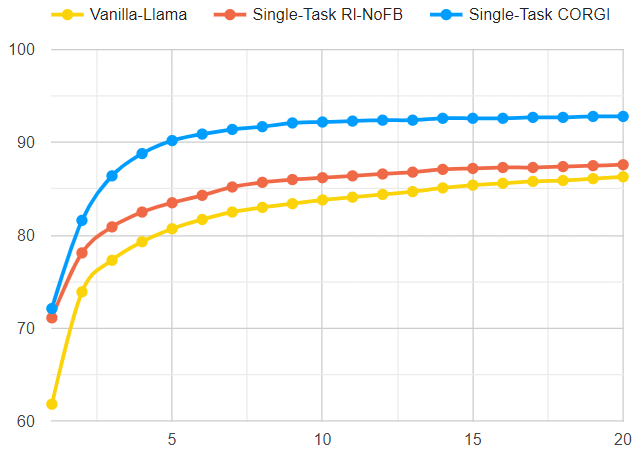}
  \caption*{ST results for the source tasks}
\end{minipage}
\begin{minipage}{.33\textwidth}
  \centering
  \includegraphics[width=1\linewidth]{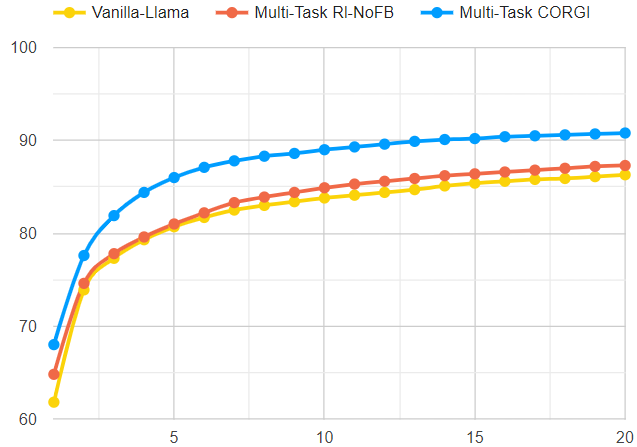}
  \caption*{MT results for source tasks}
\end{minipage}
\begin{minipage}{.33\textwidth}
  \centering
  \includegraphics[width=1\linewidth]{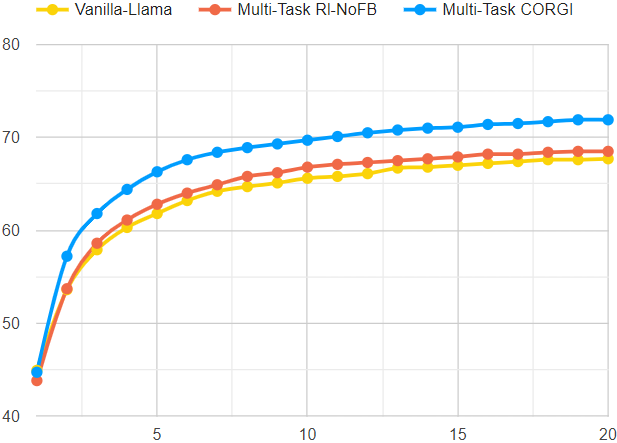}
  \caption*{MT results for the target tasks}
\end{minipage}
\captionof{figure}{ Success rates over 10 attempts for Llama-2-based models (first row) and 20 attempts for Llama-3-based models (second row) for single-task training (ST) and multi-task training (MT). The leftmost column shows results for single-task training on source tasks, the middle column displays results for multi-task training on source tasks, and the rightmost column presents meta-learning results for target tasks with multi-task models. In all settings, CORGI consistently achieves the best performance in later attempts. Additionally, the performance gap between CORGI and the baselines widens over successive attempts, highlighting CORGI's superior learning and adaptation capabilities. }
\label{lines}
\end{figure*}

\section{Results and Analysis}
We report the main results in Table \ref{tab:summarized-results} and Figure \ref{fig:llama3_specific_target_tasks_bar}. For all tasks we report the success rate metric, where success is defined as satisfying all the constraints of the task. Our experiments show that CORGI significantly improves the performance over the baselines, in both in-task and meta-learning experiments.

\subsection{Source Tasks Results using Single Task Training}
In Table \ref{tab:summarized-results} we present the source task results for the single-task trained models (task-specific results are provided in the appendix). For both Llama models, we observe that both RL-NoFB and CORGI outperform the Vanilla-Llama settings. Notably, CORGI training proves to be significantly more effective than RL-NoFB, showing an average improvement of 2\% in Llama-2 and 5\% in Llama-3. These results indicate that CORGI training effectively leverages critique feedback to substantially enhance generator performance.


Figure \ref{lines} shows result as a function of the number of critique feedback iterations. It can be seen that access to multiple critique feedbacks is beneficial for all three models, with performance further increasing after RL training. Even though the RL-NoFB model was not trained with feedback, it still benefits from these, though it did not perform as well as CORGI, which was specifically trained with this feedback. We also observe that the performance gap between baseline RL and CORGI widens with the first feedback, indicating that the CORGI model makes better use of feedback.

\subsection{Multi Task and Meta Learning Results}

\paragraph{Source Tasks Results}  The results for multi-task training on source tasks are presented in Table \ref{tab:summarized-results} (see also task-specific figures in the Appendix). Both the RL-NoFB and CORGI models show improvement compared to the Vanilla-Llama setting, with CORGI consistently outperforming RL-NoFB.

For Llama-2, multi-task training yields results comparable to single-task training, with a slight improvement in the CORGI setting. Conversely, in Llama-3, single-task training appears to be more effective than multi-task training in the CORGI setting. The limited improvement of multi-task training on the source tasks is expected, as it may be more beneficial for the model to focus specifically on the tasks it is being evaluated on.

Figure \ref{lines} shows performance over multiple feedback iterations. Initially, RL-NoFB performs comparably to (Llama-3) or outperforms (Llama-2) CORGI due to its more greedy approach. However, as the attempts progress, CORGI improves more rapidly, widening the performance gap between the two methods. This gap-widening behavior is also evident in the Llama-3 setting. These results indicate that CORGI benefits more from iterative feedback.

\paragraph{Target Tasks Results}
We next discuss the results when applying models to tasks they were not trained on. Here one might expect that training on multiple tasks would improve performance, which we indeed observe.

 Table \ref{tab:summarized-results} (see also Figures in the Appendix) show that multi-task CORGI models outperform the single-task ones, suggesting that multi-task training improves meta-learning. For example, multi-task CORGI improves over single-task by four percent (from $54.5$ to $58.5$) for Llama-2, and by $1.2\%$ for Llama-3.
 For the RL-NoFB model, there is no improvement for multi-task in Llama-2 and $0.9\%$ improvement for Llama-3. As illustrated in Figure \ref{lines}, Multi-Task CORGI adapts well to unseen feedback, increasing the performance gap between itself and the baseline model over multiple attempts.
 

\subsection{Ablation Study}
We conducted an ablation study to evaluate the significance of feedback information. Specifically, we tested what happens when one replaces the feedback with binary information, removing any useful details (such as the number of words in the generated output or its last word). Specifically, for imperfect outputs we use the feedback ``Your output is incorrect. Please try again.''. We use this feedback in both train and evaluation.

\begin{figure}[t]
  \includegraphics[width=0.5\linewidth]{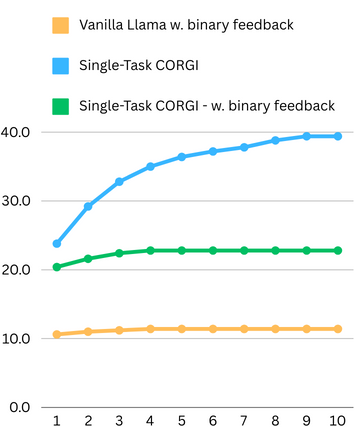}
  \centering
  \caption{Ablation experiment studying the use of binary feedback. The figure shows a Vanilla Llama model and a CORGI model that use only binary feedback in interaction (green). Also shown in  blue is the CORGI model that uses full feedback.}
  \label{fig:ablation}
\end{figure}

Figure \ref{fig:ablation} shows the results of this change, when training a Llama-2 CORGI model on the numerical planning task. It can be seen that the model with binary feedback performs considerably worse than the one with more detailed feedback. Additionally, the CORGI model still improves considerably over the Vanilla Llama baseline with binary feedback, indicating that the former has learned to use the binary feedback.  



\section{Related Work}

\paragraph{Controlled generation} There are various approaches that tackle controlled generation. Some approaches \citep{anderson-etal-2017-guided,hokamp-liu-2017-lexically,post-vilar-2018-fast,lu-etal-2021-neurologic,lu-etal-2022-neurologic} proposed constrained search algorithms that limit the lexical constraints by altering the search space. These approaches are limited to lexical constraints. Another more flexible approach is score-based sampling, where constraints are represented via a differentiable score function. For instance, \citet{liu-etal-2022-dont} represented lexical constraints via a differentiable n-gram matching function.

A more closely related approach to our method is \citet{han2023pive}, where a verification module is trained to provide feedback to a generation model, aiding in the refinement of graph-based tasks. In contrast, our approach involves training the generator LLM while incorporating feedback, whereas the generator LLM in \citet{han2023pive} remains frozen without additional training. showcases the effectiveness of meta-learning in controlled generation. This approach fine-tunes a model on multiple controlled generation tasks, allowing it to adapt to new constraints without retraining. In our work, we take this concept further by creating an interactive environment where the model receives real-time feedback from a critique system, allowing iterative refinements, leading to improved generalization when encountering new constraints.

\paragraph{Reasoning with Feedback Loops} Feedback loops are increasingly used to allow models to refine their outputs. \citet{akyurek2023rl4f} trains a critique mechanism to automate feedback provided to the generator model, aiding in its refinement process. \citet{paul2023refiner} fine-tunes both critique and generator models. In this method, the generator is trained to produce intermediate reasoning steps, which are then evaluated by the critic model, facilitating a continuous improvement process. This method required annotated data for the intermediate steps for training both the critique and generator models. \citet{besta2024graph} and \citet{long2023large} introduce the GoT and ToT frameworks, which integrate feedback mechanisms into the generation process, enhancing the COT process. \citet{miao2023selfcheck} allows LLMs to examine and evaluate their reasoning steps. \citet{chen2023interact} uses different language models  as auxiliary roles, such as checkers and sorters, to help the main language model avoid erroneous and inefficient actions. \citet{yao2022react} proposes constructing prompts using thought-act-observation triplets aiming to enhance reasoning, planning, and interaction with the environment. \citet{shinn2024reflexion} uses a dynamic feedback system, capturing recent feedback and persistent long-term storage to retain key insights, providing a robust mechanism for ongoing refinement. Inspired by these approaches, our work aims to enhance feedback utilization in the LLM refinement process through reinforcement learning. In addition, our approach does not require labeled data and allows for multiple generations during training.

\paragraph{Alignment with reinforcement learning} Reinforcement learning techniques play a critical role in optimizing models through feedback-driven reward signals. \citet{ouyang2022training} and \citet{lee2023rlaif} reinforce a supervised fine-tuned (SFT) model. Specifically, they used a PPO approach for aligning the model with human feedback by training a reward module based on human preferences. \citet{lee2023rlaif} replace human preference labels with those of the model itself. The model is prompted to evaluate its own predictions post-supervision, in consideration of human values and preferences. 
These approaches primarily focus on refining models through explicit reward signals. Our method integrates both textual critique and reinforcement learning by incorporating a conversational critique mechanism which complements the reinforcement-based rewards.

\section{Conclusion}
We present an approach for improving the capability of LLMs to use feedback in an interactive manner. Our key observation is that a model can learn to use such interactive feedback via reinforcement learning. A key advantage of our scheme is that it does not require labeled data (i.e., examples of correct generation). Instead it just employs a critique which can score generations.

Our results show that models indeed benefit from this training scheme. Not only can they improve on tasks that were used during RL, but they also improve on very different tasks (e.g., clustering and Panagram) that they were not trained on. This suggests that CORGI training indeed endowed the model with a capability to receive a textual critique in an interactive fashion, and continuously improve ``on tape'' using this critique.

Our work raises several questions for followup. One is the mechanism via which the model learns to use the reward. Namely, how does the textual feedback translate into the action of improving generation. It will be interesting to study this using tools recently employed in understanding in-context learning \citep[e.g.,][]{hendel-etal-2023-context,function_vectors}. A related question is
optimizing over the textual feedback. Namely, is it possible to effectively optimize over the specific text that is used in feedback (e.g., similar to soft-prompt tuning 
\citet{lester2021power}).

Our approach can be used to significantly enhance the quality of controlled generation of text, thus improving the applicability of LLMs in domains such as assistants and education. 

Further research could explore using human input in addition to the critique input to better adjust the LLM to user preferences. It will also be interesting to investigate more algorithmic tasks where critiques not only provide scores but also suggest strategies for improvement. Exploring these directions may yield a richer understanding of how models can learn and adapt in more complex, real-world scenarios.


\bibliography{teaching_models_to_Improve_on_tape}

\appendix

\section*{Data Construction} \label{appendix:data_construction}

In this section, we outline the tasks, along with the methods used for reward calculation and feedback generation. Detailed examples are provided in Figures \ref{fig:source_tasks_examples}, \ref{fig:target_tasks_examples}, \ref{fig:target_tasks_llama_examples}.

\paragraph{Sentiment Reviews Generation} 
\begin{itemize}
    \item \underline{Task:} As described in \citet{sun-etal-2023-evaluating} the generation task is to generate text about a given product name, with a specified star rating. The objective is to generate a review for the product corresponding to the given star rating.
    \item \underline{Data Construction:} To create the dataset, we utilize the Amazon US Customer Reviews Dataset to obtain product names and randomly assign star ratings. The critique is constructed by using a classifier that was trained on this dataset \footnote{\url{https://huggingface.co/LiYuan/amazon-review-sentiment-analysis}}.
    \item \underline{Critique Feedback:} If the output is incorrect, the critique feedback informs the generator of the actual number of stars that were classified. 
     \item \underline{Critique Scoring:} The scoring for this output is binary: it receives a 1 if the generated review matches the specified number of stars, and 0 if it does not.
\end{itemize}

\paragraph{Story Generation}  
\begin{itemize}
    \item \underline{Task:}  As described in \citet{sun-etal-2023-evaluating} the generation task is complete a beginning of a story with four sentences, ensuring diversity and coherence. 
    \item \underline{Data Construction:} The dataset utilized for this task is ROCStories \citep{mostafazadeh-etal-2016-corpus}. We use the first sentence of each story as the given prefix.
    \item \underline{Critique Feedback:} The critique is constructed by using two components: a repetition component that validates the rep-4 metric \citep{welleck2019neural}  which measures the proportion of duplicate tokens within a window of 4 in the generated output, and a coherence validation by thresholding the cosine similarity between the embedding of the story prefix and its continuation, as facilitated by SimCSE \citep{gao-etal-2021-simcse}. If the output is incorrect, the critique feedback informs the generator about repeated tokens or whether the story is incoherent. 
    \item \underline{Critique Scoring:} The scoring is the average of two scores: the repetition score and the coherence score. The repetition score is 1 if the rep-4 metric is 0 (indicating no repetition), otherwise, it is 0. The coherence score is 1 if the cosine similarity exceeds 0.3, otherwise, it is 0.
\end{itemize}

\paragraph{Rationale Generation} 
\begin{itemize}
    \item \underline{Task:} As described in \citet{sun-etal-2023-evaluating} the generation task is to generate contextual information for a multiple choice common sense question, such that another model can accurately answer the question. We refer to the model responsible for answering the question and measure the contextual information quality as the \textit{reader model}. Example can be seen in Figure \ref{fig:source_tasks_examples}.
    \item \underline{Data Construction:}  The questions are taken from ECQA \citep{aggarwal-etal-2021-explanations} and employed FlanT5-Large \citep{chung2024scaling} as the reader model.
    \item \underline{Critique Feedback:} The critique is constructed based on the reader model's performance. If the output is incorrect, the critique informs the LLM that the question was answered incorrectly.
    \item \underline{Critique Scoring:} The scoring is decided as follows: if the reader correctly answers the question, the reward is 1; otherwise, it is 0.
    
\end{itemize}

\paragraph{Numerical Planning} 
\begin{itemize}
\item \underline{Task:} As described in \citet{sun-etal-2023-evaluating}, the generation task is to complete a given prefix with a specified number of words, $N$, and to conclude with the specified last word, $w$.
\item \underline{Data Construction:} The prefixes are generated from the Reddit Short Stories dataset\footnote{\url{https://huggingface.co/datasets/AlekseyKorshuk/romancebooks}, \url{https://www.kaggle.com/datasets/trevordu/reddit-short-stories}}. We sample sentences from the stories and randomize a number $N$ between 2 and 10 (or the sentence length, whichever is shorter). We remove the last $N$ words from each sentence to form the prefix, and $N$ is the given number. The final word of the sentence is the requested last word.
\item \underline{Critique Scoring:} The scoring for this task is a weighted average of two scores: the length score and the last word score. The length score is 1 if the generated output has the specified number of words, otherwise 0. The last word score is 1 if the final word of the generated sentence is the specified word, otherwise 0. Since matching the length is more challenging for the generator than matching the last word, we set the weight of the length score to 0.8 and the weight of the last word score to 0.2.
\end{itemize}

\paragraph{Controlled Paraphrase Generation} 
\begin{itemize}
\item \underline{Task:} As described in \citet{sun-etal-2023-evaluating}, the generation task is to create a paraphrase of the query that matches the syntax structure of the exemplar. 
\item \underline{Data Construction:} The queries are taken from the QQP-Pos dataset \citep{kumar-etal-2020-syntax}. We construct the exemplars by generating the parsing tree for the paraphrase of each query.
\item \underline{Critique Feedback:} The critique is constructed by confirming that the generated output is a valid paraphrase using a paraphrase classifier \citep{nighojkar-licato-2021-improving} and by verifying the syntax structure with the Stanford CoreNLP parser. We used the QQP-Pos dataset. Example can be seen in Figure \ref{fig:source_tasks_examples}. If the output is incorrect, the critique informs the LLM the generated output's current syntax parse and whether it is classified as a paraphrase. \item \underline{Critique Scoring:} The scoring for this task is a combination of two scores: the paraphrase score and the structure score. The paraphrase score is 1 if the generated query is classified as a paraphrase of the original query, otherwise 0. The structure score is 1 if the generated query matches the exact requested structure, otherwise 0. The final scoring is the average of these two scores.

\end{itemize}

\paragraph{Style Transfer} 
\begin{itemize}
\item \underline{Task:} We define the style transfer task, where the generation task is converting biased text into neutral text.

\item \underline{Data Construction:} We used the Wiki Neutrality Corpus \citep{pryzant2020automatically}, by taking the biased texts. 
\item \underline{Critique Feedback:} The critique is constructed by using a classifier trained on this dataset\footnote{\url{https://huggingface.co/cffl/bert-base-styleclassification-subjective-neutral}}. If the output is incorrect, the critique informs the LLM that the text is classified as biased.
\item \underline{Critique's Scoring:} The generator receives a score of 1 if the generated text is classified as neutral and 0 otherwise.
\end{itemize}

\paragraph{Clustering} 
\begin{itemize}
\item \underline{Task:} The generator's input is a list of student names, and the generation task is to group the students into clusters of two or more, considering their preferences. These preferences specify which students they do or do not want to be grouped with.
\item \underline{Data Construction:} We sample 4 to 10 student names. The number of constraints is a random value between 1 and half of the total number of students. We ensure the constraints can be satisfied by generating them based on a predefined random grouping.
\item \underline{Critique Feedback:} The critique is constructed by ensuring that each student is correctly grouped, meaning they are placed in a group of at least two and their preferences are met. If the output is incorrect, the critique informs the LLM about the specific constraints that were violated.
\item \underline{Critique's Scoring:} The scoring is defined by the percentage of students correctly grouped.
\end{itemize}

\paragraph{Panagram} 
\begin{itemize}
\item \underline{Task:} In this task, the generation task is to generate a word using a given list of letters. Specifically, the model must create a word that incorporates all the specified letters, with each letter being used at least once and possibly multiple times.
\item \underline{Data Construction:} We sample words containing up to six letters. The task prompt is then constructed using the set of letters from the sampled word.
\item \underline{Critique Feedback:}  The critique is constructed by ensuring that the generated word is a valid English word, contains no extra letters, and includes all the given letters. If the output is incorrect, the critique informs the LLM about the specific constraints that were violated.
\item \underline{Critique's Scoring:}  The scoring is 0 if the word is not valid. Otherwise, it is determined by the percentage of the specified letters used correctly.
\end{itemize}

\paragraph{CommonGen-lite} 
\begin{itemize}
\item \underline{Task:} The generation task is to construct a sentence given a list of words and their corresponding parts of speech. The sentence must ensure that each word aligns with its designated part of speech and collectively describes a common scenario. 
\item \underline{Data Construction:} For the sentences and the concept set (i.e., the words and their parts of speech), we used data from the CommonGen-lite dataset\footnote{\url{https://github.com/allenai/CommonGen-Eval}}.
\item \underline{Critique Feedback:} The critique is constructed by ensuring that each word was used with its correct part of speech using Spacy parser \citep{spacy2}. 
Additionally, the critique verifies that the generated text describes a common-sense scenario. Instead of comparing against a reference sentence, which the critique does not have access to, we employ Gemma-1.1 \citep{team2023gemini}. This model determines whether the generated output depicts a common-sense scenario. If it does not, Gemma-1.1 provides an explanation as feedback to the generator. 
\item \underline{Critique's Scoring:} The scoring consists of three components: cover, representing the percentage of concepts used; PoS (Part of Speech), indicating the percentage of concepts where the correct part of speech is applied; and a common sense score, which is 1 if Gemma-1.1 confirms the scenario as common sense and 0 otherwise. The final score is the average of the cover score, the PoS score, and the common sense score.
\end{itemize}

\paragraph{Program Synthesis} 
\begin{itemize}
\item \underline{Task:} This task is adapted from \citet{srivastava2023beyond}. The goal of the generator is to generate the simplest Python function that accurately maps a given set of input-output pairs.
\item \underline{Data Construction:} We utilized the input-output mappings provided by \citet{srivastava2023beyond}. We filtered the dataset to include only functions from the numerical category, as Llama-3 demonstrated poor performance on other categories.
\item \underline{Critique Feedback:} The critique provides feedback to the generator on two key aspects: whether the function's syntax is correct and which specific input-output pairs the function failed to map correctly.
\item \underline{Critique Scoring:} The score is assigned as follows: if the function's syntax is invalid, the score is 0. Otherwise, the score is based on the percentage of correct input-output mappings.
\end{itemize}

\paragraph{MBPP}
\begin{itemize}
\item \underline{Task Description:} This task is adapted from \citet{austin2021program}. The objective of the generator is to create a Python function that fulfills the given instruction.
\item \underline{Data Construction:} We used the sanitized subset of data provided by \citet{austin2021program}.
\item \underline{Critique Feedback:} The critique provides feedback on two main aspects: the correctness of the function's syntax and the specific unit tests that the generated function fails.
\item \underline{Critique Scoring:} The scoring is as follows: if the function's syntax is invalid, the score is 0. Otherwise, the score is determined by the percentage of unit tests that pass.
\end{itemize}

\paragraph{CommonGen Hard}
\begin{itemize}
\item \underline{Task Description:} This task, adapted from \citet{NEURIPS2023_91edff07}, challenges the generator to produce a sentence that incorporates all of the provided keywords, which range between 20 to 30 in total.
\item \underline{Data Construction:} We utilized the dataset provided by \citet{NEURIPS2023_91edff07}.
\item \underline{Critique Feedback:} The critique highlights any keywords that are missing from the generated sentence.
\item \underline{Critique Scoring:} The score is based on keyword coverage—the percentage of the given keywords successfully included in the sentence.
\end{itemize}

\section{Limitations}
We identify several limitations of the CORGI method. Firstly, its effectiveness is constrained by the context length, as the feedback it provides is limited to this span. Secondly, it necessitates training and adjusting the weights of the language model, making it impractical for larger models such as Llama-2-70b-chat and Llama-3-70b-instruct or commercial models like GPT4. Finally, we focus here on tasks where good critiques are available. However, for tasks such as creative writing, critiques are likely to be suboptimal.

\section*{Task-Specific Results}
The following figures display the success rates of CORGI and the baseline models on both source and target tasks, organized by task splits and base models. Each chart includes results from both Single-Task and Multi-Task settings, highlighting the effectiveness of CORGI training as well as its meta-learning capabilities. \\

\begin{figure}[h]
  \includegraphics[width=1\linewidth]{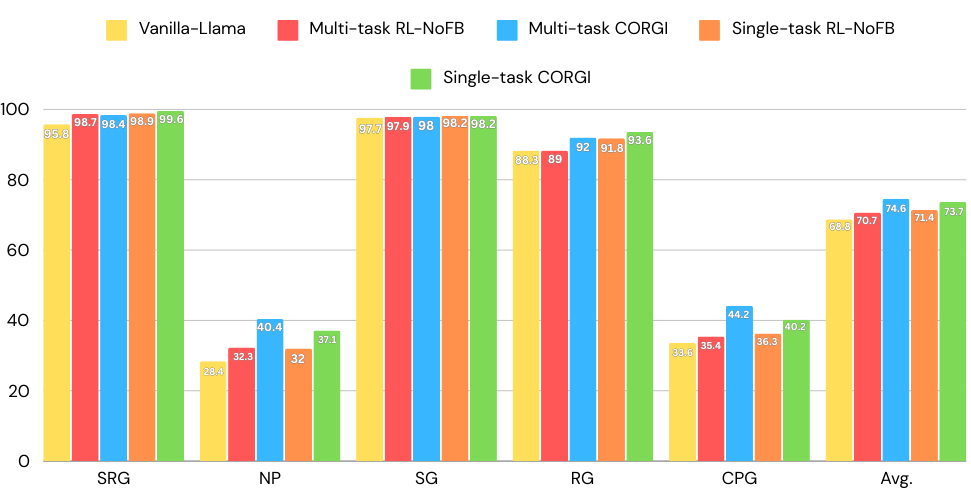}
  \centering
  \caption[Llama-2 results on source tasks, split by task]{ Llama-2 results on source tasks, when split by task, show that both RL-NoFB and CORGI training outperform the Vanilla-Llama baseline. Notably, CORGI training consistently surpasses both the RL-NoFB and Vanilla-Llama baselines across nearly all tasks in both single-task and multi-task training settings. }
  \label{fig:source_tasks_llama_2_bar_results}
\end{figure}

\begin{figure}[h]
  \includegraphics[width=1\linewidth]{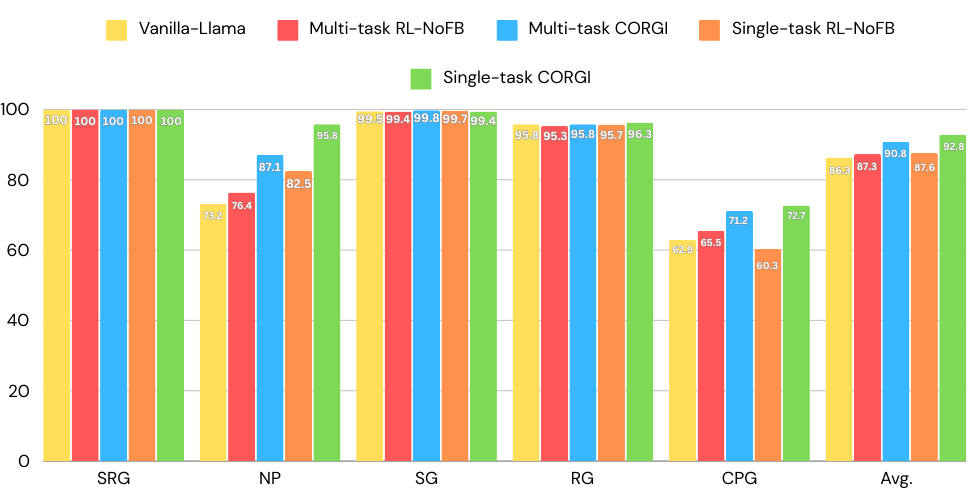}
  \centering
  \caption[Llama-3 results on source tasks, split by task]{ The Llama-3 results on source tasks, when split by task, demonstrate that CORGI training outperforms both Vanilla-Llama and RL-NoFB training for tasks where Vanilla-Llama doesn't achieve nearly perfect results. RL-NoFB training is only effective for the numerical planning task. Additionally, single-task training appears to be more effective than multi-task training. }
  \label{fig:source_tasks_llama3_bar_results}
\end{figure}

\begin{figure}[H]
  \includegraphics[width=1\linewidth]{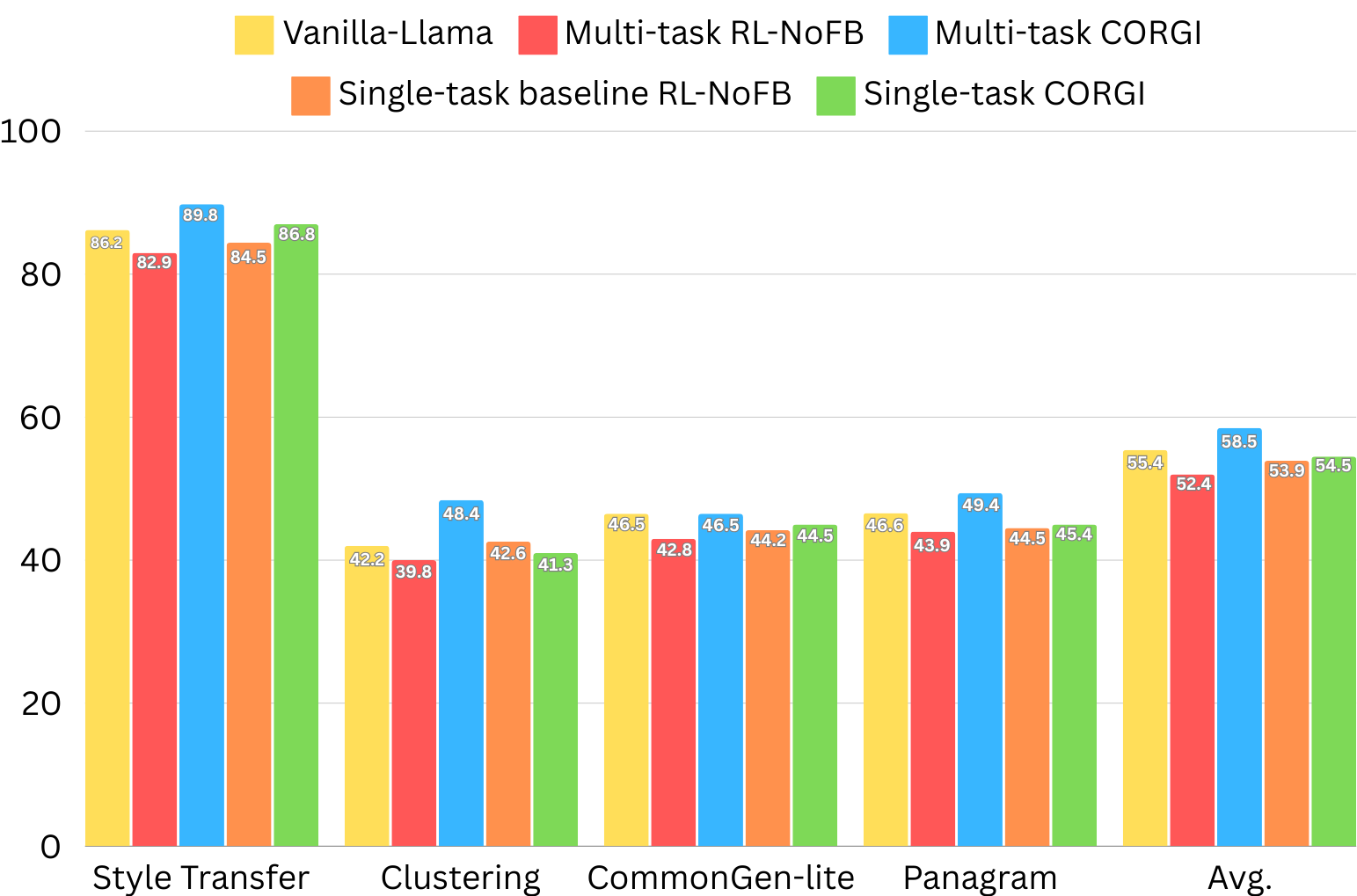}
  \centering
  \caption[Llama-2 results on target tasks, split by task]{ Llama-2 results on target tasks, when split by task. The effectiveness of meta-learning is particularly evident in the target tasks, where the multi-task CORGI model, which did not encounter these tasks during training, outperforms the baselines. On the other hand we can see that the multi-task RL-NoFB doesn't have the capacity to generalize across rewards without being provided with feedback during training. }
  \label{fig:target_tasks_llama2_bar_results}
\end{figure}

\begin{figure}[H]
  \includegraphics[width=1\linewidth]{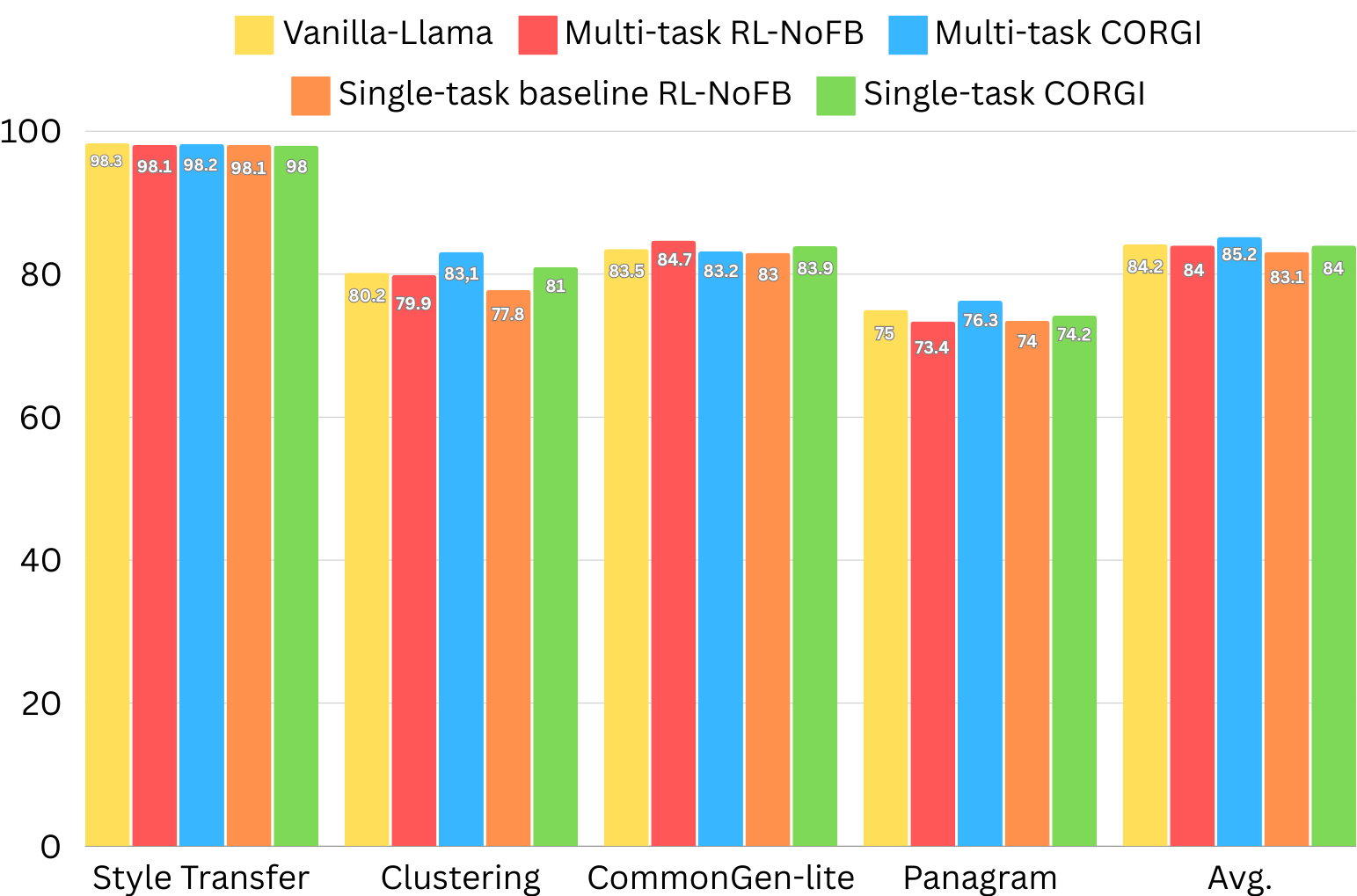}
  \centering
  \caption{ Llama-3 results on target tasks, split by task (excluding the Llama-3 specific tasks which can be seen in Figure 2 in the main paper). The RL-NoFB model shows comparable performance to the Vanilla-Llama. CORGI training, however, achieves superior results on both the Clustering and Panagram tasks. }
  \label{fig:target_tasks_llama3_bar_results}
\end{figure}

\section*{Output Extraction}

Some challenges in this setup include handling long sequences during training, which require a lot of memory, and extracting the relevant answers from the output. These models, not being fine-tuned specifically for such tasks, tend to generate verbose outputs, making it difficult to extract the relevant information. For example, the model might start with a general assistant statement or apologize for a mistake when given feedback, before generating the requested output.

To overcome this behavior, we modify the chat template used by the chat models. Typically, models fine-tuned on chat tasks use a template where the task is enclosed with start and end tokens. Even when given few-shot examples—where the model is shown a few pairs of inputs and outputs to illustrate the desired format, like "input: output:"—the model might still produce unnecessary text. The original instruction template is as follows:

\begin{flushleft}

\path|<user_start_token> <instruction_tokens> Output: <user_end_token>|

\end{flushleft}

We change this template to:

\begin{flushleft}

\path|<user_start_token> <instruction_tokens> <user_end_token> Output:|

\end{flushleft}

This change prompts the model to generate the relevant text directly. We apply this change to both the initial instruction and the feedback. Although this template differs from the one used during fine-tuning, we find that this method is not only more efficient in terms of memory usage and answer extraction but also improves performance, as it leverages the pretraining of the language model.

\section*{Hyper-Parameters}
We used the Adam optimizer with a learning rate of $10^{-5}$ and Adaptive KL control \cite{ziegler2019fine}, setting the initial KL coefficients at $0.05$ for Llama-2 and $0.075$ for Llama-3. To address training instability observed in the rationale generation task for Llama-3, we increased the KL coefficient to $0.3$. We experimented with KL divergence values ranging from $0.025$ to $0.3$, selecting the optimal setting based on performance on the validation set.

Additionally, we set $\gamma = 1$, $\lambda = 0.95$, and $\epsilon = 0.2$. The training batch size was set to $32$, with $4$ optimization epochs per batch.

\section*{Examples} \label{appendix:examples}
\begin{figure*}[h]
  \includegraphics[width=0.7\linewidth]{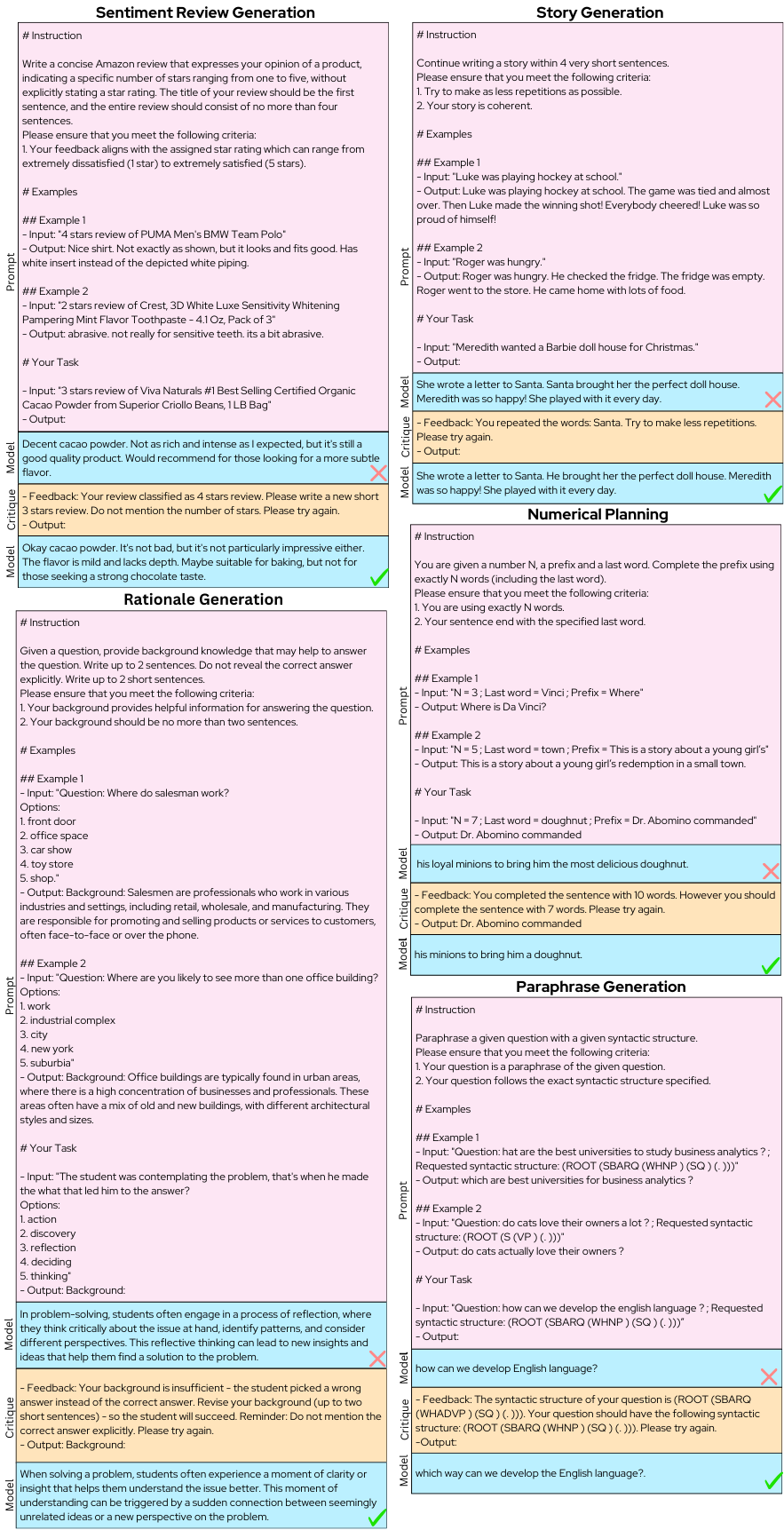}
  \centering
  \caption{ Examples of conversations between the LLM and the critique regarding the source tasks. }
  \label{fig:source_tasks_examples}
\end{figure*}

\begin{figure*}[h]
  \includegraphics[width=0.8\linewidth]{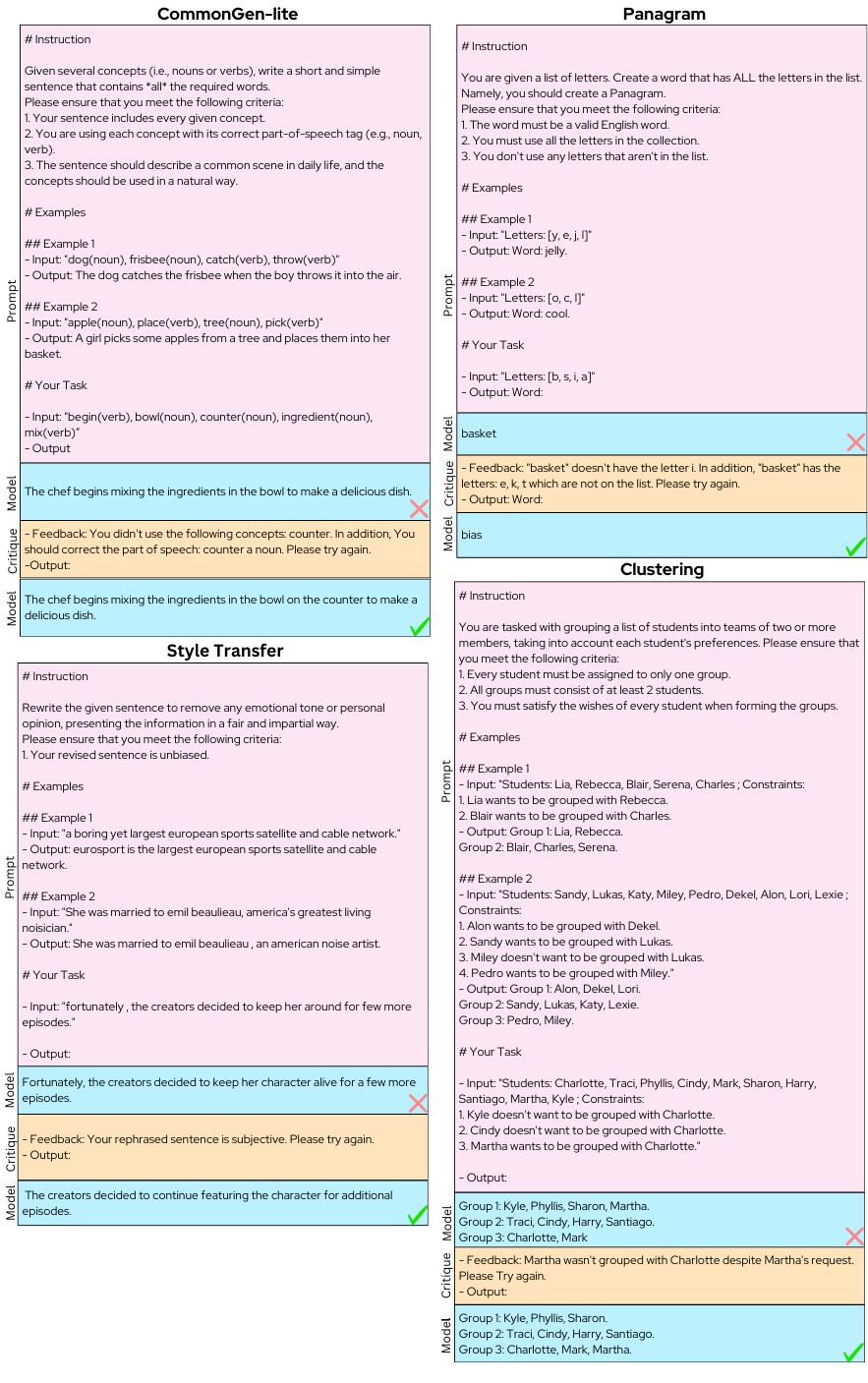}
  \centering
  \caption{ Examples of conversations between the LLM and the critique regarding the target tasks for both Llama-2 and Llama-3 models. }
  \label{fig:target_tasks_examples}
\end{figure*}

\begin{figure*}[h]
  \includegraphics[width=0.8\linewidth]{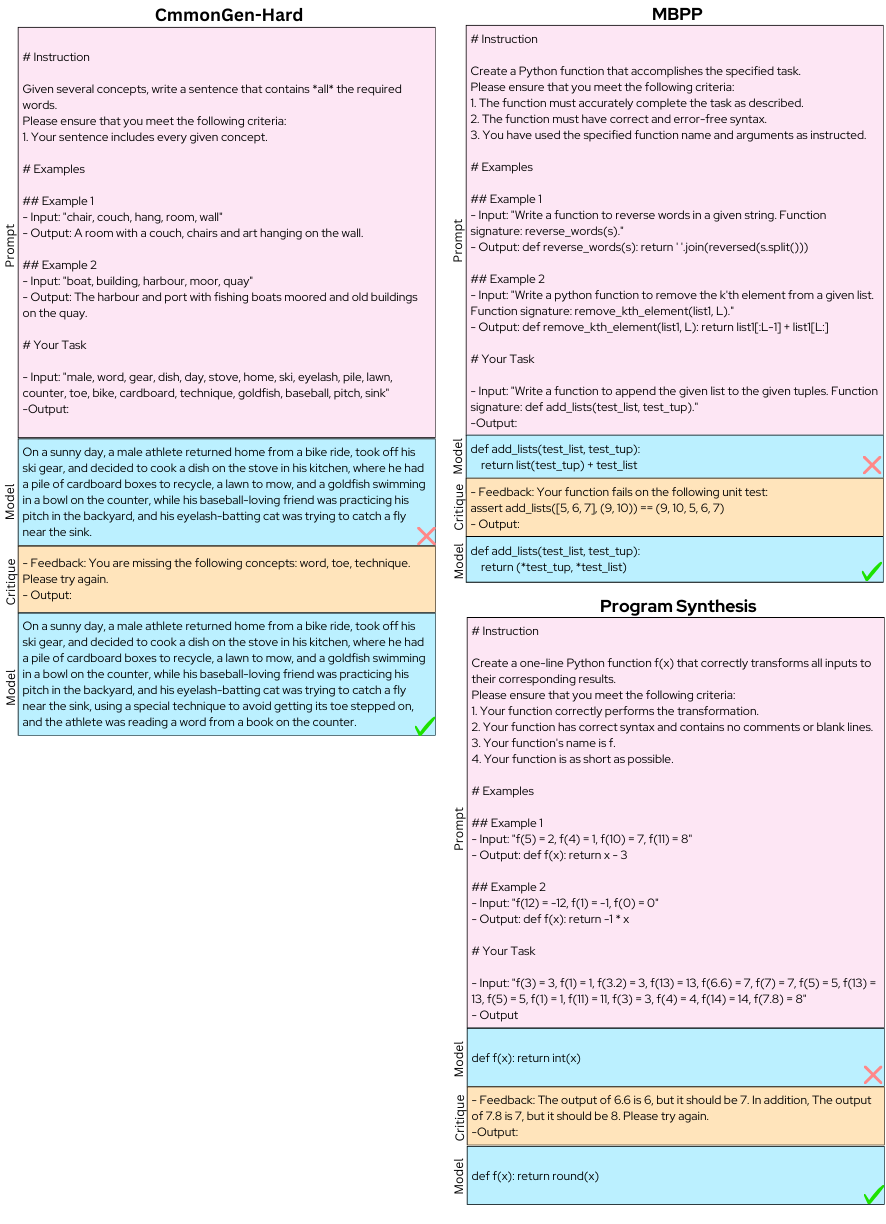}
  \centering
  \caption{ Examples of conversations between the LLM and the critique regarding the Llama-3-specific target tasks. }
  \label{fig:target_tasks_llama_examples}
\end{figure*}

Examples for source tasks, Llama-2 and Llama-3 target tasks and Llama-3-specific target tasks can be seen in Figures \ref{fig:source_tasks_examples}, \ref{fig:target_tasks_examples}, \ref{fig:target_tasks_llama_examples} respectively.

\end{document}